\documentclass[a4paper]{article} 
 
\usepackage{graphicx}
\usepackage{mathptmx}     
\usepackage{amsmath}
\usepackage{times}
\usepackage{color}
\usepackage{multirow}
\usepackage[authoryear]{natbib}
\usepackage{rotating}
\usepackage{bbm}
\usepackage{latexsym}
\usepackage{enumerate}

\begin{document}

\title{A Note on Topology Preservation in Classification, and the Construction of a Universal Neuron Grid}

\author{Dietmar Volz\\
\ttfamily dietmar.volz@gmx.net}

\maketitle

\begin{abstract}
It will be shown that according to theorems of K. Menger, every neuron grid if 
identified with a curve is able to preserve the adopted qualitative structure of a data space. 
Furthermore, if this identification is made, the neuron grid structure can always be mapped to a subset of 
a universal neuron grid which is constructable in three space dimensions. Conclusions will be drawn for established neuron grid types as well as neural fields.
\end{abstract}

\section{Mathematical Preliminaries}
\label{MP}
Topology is one of the basic branches of mathematics. It is sometimes also referred to as qualitative geometry, 
in a way that it deals with the qualitative properties and structure of geometrical objects. The geometrical 
objects of interest in this paper are vector spaces, manifolds, and curves. These form the basis of the presented 
mathematical treatment of clustering with neuron grids. Consequently, the paper has to begin with some mathematical preliminaries.  
\subsection{Manifolds}
\label{MAN}
\begin{figure}[h]
\hfill
\begin{center}
\includegraphics[width=2.5in]{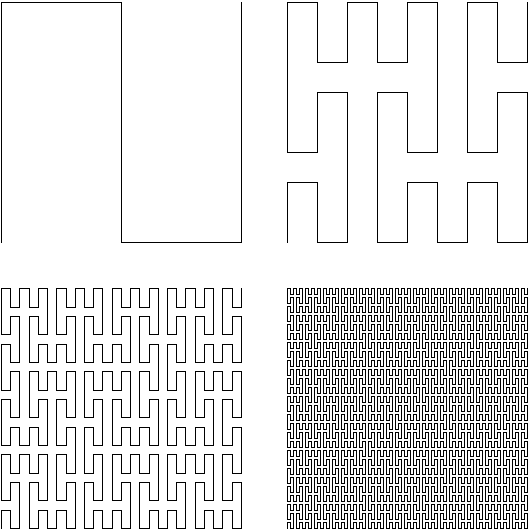}
\end{center}
\caption{A Peano curve (source: Wikipedia)}
\label{Fig: A Peano curve (source: Wikipedia)}
\end{figure}

In the following, a n-dimensional vector space will be identified with a subspace of ${\mbox{$I\!\!R$}^{n}}$. 
It is assumed that ${\mbox{$I\!\!R$}^{n}}$ is equipped with a topology which in turn is induced by a metric. 
Mappings between subspaces of ${\mbox{$I\!\!R$}^{n}}$ are called {\em homeomorphic} or {\em topology preserving} 
if they are one-to-one and continuous in both directions. It is known from the theorem of dimension 
invariance \citep{Brouwer1911} that mappings between non-empty open sets $U \subset  {\mbox{$I\!\!R$}^{m}}$ 
and $V \subset  {\mbox{$I\!\!R$}^{n}}$ for ${m\neq n}$ are never homeomorphic. In the light of Brouwer's theorem 
it is the {\bf open sets} that fix the topological dimension of a subset of ${\mbox{$I\!\!R$}^{n}}$. 
As the ${\mbox{$I\!\!R$}^{n}}$ is introduced as a metric space, the open sets are given by open 
balls that formalize the concept of distance between points of this metric space. Homeomorphic mappings 
preserve the neighborhood relationship between points of ${\mbox{$I\!\!R$}^{n}}$. Amongst the huge variety 
of subsets of the metric space ${\mbox{$I\!\!R$}^{n}}$, the n-dimensional manifolds (or n-manifolds) have 
turned out to be of interest as these describe solution spaces of equations or geometrical entities. 
Manifolds are parameterized geometrical objects, parameters could describe e.g. the coordinates of a 
physical space. As n-manifolds are locally Euclidean of dimension n they are subject to Brouwer's theorem 
and therefore of fixed topological dimension. As a consequence, dimension reducing mappings between manifolds will not be able to transfer mutual topological structures. A vivid example of the dimension conflict of two manifolds of different dimension is given by a surjective 
and continuous mapping of $[0,1]$ onto $[0,1] \times [0,1]$ which is also called a 'Peano curve' \citep{Peano1890}. 
The convoluted structure of a Peano curve does not reflect the neighborhood relationship of elements of 
the underlying manifold $\subset  {\mbox{$I\!\!R$}^{2}}$, not even locally: There is no topology preserving 
mapping of $[0,1]$ onto $[0,1] \times [0,1]$. An illustration of a Peano curve to $4^{th}$ iteration is given in figure \ref{Fig: A Peano curve (source: Wikipedia)}.\\

\subsection{Curves}
The Peano curve introduced in the preceding section is based on the conventional definition of a compact curve as a continuous 
mapping of $[0,1]$. The following definition according to \citep{Menger1968} renders the definition of a compact curve more precisely. Beforehand, a definition of a {\em continuum} is required.\\ 
\noindent
{\bf Definition} {\it A compact, connected set ${\subset \mbox{$I\!\!R$}^{n}}$
having more than one element is a continuum.}\\
A set ${\subset \mbox{$I\!\!R$}^{n}}$ which contains no continuum is called {\em discontinuous}.\\
\noindent
{\bf Definition} {\it A continuum $K$ as a subset of a metric space is called a curve if every point of $K$ is 
contained in arbitrary small neighborhoods having discontinuous intersects with $K$.}\\
From Menger's definition of a curve the following theorem results:\\
{\bf Theorem}[Menger] {\it Every curve defined in a metric space is homeomorphic
to a curve defined in ${\mbox{$I\!\!R$}^{3}}$ .}\\

\noindent
The proof of this theorem is left here. The interested reader is referred to Menger's textbook \citep{Menger1968}.
The theorem states that with regard to topological aspects, the transition from curves defined in 
three-dimensional Euclidean space to curves defined in an arbitrary metric space doesn't give any generalization.\\
Menger's definition of a curve even leads to another theorem \citep{Menger1968}.\\
{\bf Theorem}[Menger] {\it Every compact curve in a metric space can be mapped to a subset of a so-called universal curve $\subset {\mbox{$I\!\!R$}^{3}}$.}\\

\begin{figure}[h]
\hfill
\begin{center}
\includegraphics[width=3.2in]{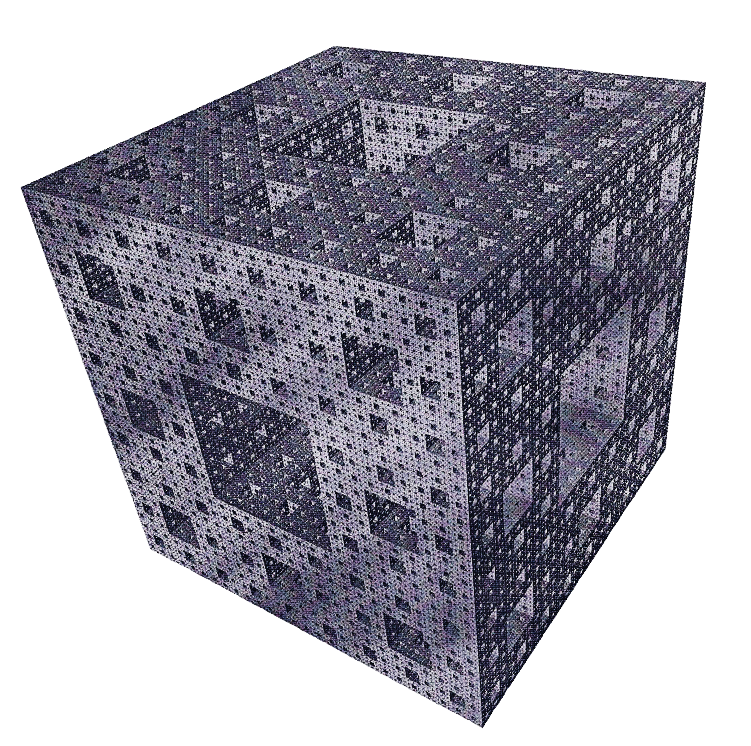}
\end{center}
\caption{Menger's universal curve in perspective view}
\label{Fig: Menger's universal curve in perspective view}
\end{figure}

\noindent
The {\em universal curve} $K$ is constructable and up to homeomorphism uniquely defined. 
The construction starts with a cube ${[0,1]}^{3} \subset {\mbox{$I\!\!R$}^{3}}$.  $K$ is the 
set of all points of ${[0,1]}^{3}$ such that at least two of the three coordinates have triadic 
expansions that do not contain a 1. The hereby constructed set fulfills the definition of a 
curve \citep{Menger1968} and leads to a self-similar structure. The construction is shown in figure \ref{Fig: Menger's universal curve in perspective view}.
 
\noindent
Summarizing, the topological relationships that arise from Brouwer's and Menger's theorems are as follows:\\

\indent
$X \subset {\mbox{$I\!\!R$}^{n}} \quad\xrightarrow{\quad\not\cong\quad}\quad Y \subset 
{\mbox{$I\!\!R$}^{m}}\quad\xrightarrow{\quad\not\cong\quad}\quad Z \subset 
{\mbox{$I\!\!R$}^{3}} \quad, \quad$  if  $\quad m\neq n$ and $m\neq 3$\\

$C \subset {\mbox{$I\!\!R$}^{n}} \quad\xrightarrow{\quad\qquad\qquad\cong\qquad\qquad\qquad}\quad C \subset {\mbox{$I\!\!R$}^{3}} \quad, \quad \forall n$\\

hereby X, Y, Z are (n,m,3)-manifolds, $C$ a curve, $\cong$ identifies a homeomorphic mapping.\\
Subsets of the universal curve as introduced in this section usually are no manifolds. 
The only exception are the trivial cases of subsets that are homeomorphic to an interval or $S^1$.

\section{Application to Neuron Grids}
\label{ANG}
\subsection{Vector Quantization}
\label{VQ}
The method of Vector Quantization (VQ) \citep{Linde1980} classifies elements of a 
vector space $V \subset  {\mbox{$I\!\!R$}^{n}}$ by approximating the probability density 
function $p(v)$ of elements $v$ of $V$. Hereby, the elements of $V$ are mapped iteratively to a 
set of weight vectors $U$ such that respective a suitable metric the quantization error 
functional gets minimal. An extension of the VQ method introduces a topology on the set 
of weight vectors such that $U$ constitutes a topological space which is then called a {\em neuron grid}. 
By means of the topology of the neuron grid, elements of $U$ are mutually adjacent if they are adjacent 
on the neuron grid. Using algorithms of machine learning, for example Kohonen's feature maps \citep{Kohonen1990} 
and several variants thereof \citep{Fritzke1992, Oja1999} try to transfer the topology of the vector space $V$ 
to the topology of the neuron grid $U$. Hereby, the topology of the neuron grid $U$ as adapted to the 
vector space topology $V$ is representative of the topology of $V$. The following section will provide some 
insight to the restrictions of topology preservation in clustering of a vector space with VQ methods.

\subsection{Identification of a Neuron Grid with a Curve}
\label{ING}
Assuming Menger's curve definition, the following identification is made:

\noindent 
{\bf Definition} {\it A neuron grid is a curve}

\noindent
It is required that neurons of a neuron grid constitute a discrete point set of a curve. Subsets of a neuron grid which contain no neurons are called {\em links} between neurons.\\
Consequently, the topology of the defined neuron grid corresponds to a curve topology. The identification of a neuron grid with a curve as motivated in this section is coherent and straightforward for 
neuron grids whose topologies are identifiable with a subset of Menger's universal curve. This is the case for neuron grids of the Kohonen type \citep{Kohonen1990} as well as some variants thereof e.g. \citep{Fritzke1992,Oja1999} that 
constitute topological spaces of a discrete point set together with a set of links between the point set. 
The presented identification with a curve produces the following results:

\begin{enumerate}[i]
\item All neuron grid topologies are uniquely definable in ${\mbox{$I\!\!R$}^{3}}$. There's no need to introduce 'hypercubes' or 'high-dimensional' grid topologies.
\item  Given the neuron grid topology is adapted to a vector space topology $\subset {\mbox{$I\!\!R$}^{n}}$, then 
the neuron grid topology will generally be retained in ${\mbox{$I\!\!R$}^{3}}$. Clustering results are always visualizable in ${\mbox{$I\!\!R$}^{3}}$.
\item Up to homeomorphism it exists a uniquely defined {\em universal neuron grid} such that every connected neuron grid is a subset of the universal neuron grid.
\end{enumerate}

\noindent
These results were inferred directly from Menger's curve theorems.\\ It should be mentioned though that 
Brouwer's theorem precludes a direct topology preserving mapping between the vector space topology $V$ 
which is induced by the metric of ${\mbox{$I\!\!R$}^{n}}$ and $U \subset {\mbox{$I\!\!R$}^{m}}$ 
if $m \neq n$. The conflict of dimension as illustrated in the example of section \ref{MAN} is hereby reproduced, this time as the lack of homeomorphic mappings between open balls of metric spaces. In consequence, metrical properties of ${\mbox{$I\!\!R$}^{n}}$ generally 
won't be reproduced on the clustering result. \\
It should also be mentioned that similar results are not possible for a neuron grid if it is 
defined by a two- or higher-dimensional manifold. The well-known 'Klein bottle' \citep{Klein1923} provides 
a vivid example what might happen to a two-dimensional surface that is defined in ${\mbox{$I\!\!R$}^{n}}, n>3$, and shows an apparent self-intersection if mapped to ${\mbox{$I\!\!R$}^{3}}$. The self-intersection generates a subset of points on the Klein bottle that are separate in ${\mbox{$I\!\!R$}^{n}}, n>3$ and get identified through the mapping to  ${\mbox{$I\!\!R$}^{3}}$. Hereby, two separate clusters are spuriously merged into one.\\

\section*{Discussion}
The presented theorems of Brouwer and Menger illustrate 
an alternative though simple and straightforward approach to the topological foundation of neuron grids. The identification of a neuron grid with a subset of the universal curve at a first or second glance may be seen as trivial. This is in particular due to the presented definition of a curve which is intuitive and comprehensible, also from a naive standpoint. The presented topological framework of neuron grids restricts neuron sets to discrete point sets. Alternative approaches \citep{Amari1977,Bressloff2005,Cottet1995} introduce {\em continuous} models of neural fields. As neural fields per definition provide open sets in their domain, the present paper also incloses the caveat that neural fields might produce unusable results if applied to clustering tasks.\\ 

\section*{Acknowledgements}
  I would like to thank Catalin Dartu for the nice visualization of Menger's universal curve.

\end{document}